%% file: arxiv.tex
\documentclass[letterpaper]{article} 
\usepackage[preprint]{aaai2027}
\usepackage[hyphens]{url}  
\usepackage{graphicx}  
\usepackage{natbib}  
\usepackage{caption}  
\usepackage{amsmath,amssymb}
\usepackage{booktabs}

\newcommand{\actprimary}[1]{\textbf{#1}}
\newcommand{\actsecondary}[1]{#1}

\input{author_info}

\title{S$^2$-HWM: Sparse Event-Structured Hierarchical World Model for Long-Horizon Surgical Robot Manipulation}

\begin{document}

\maketitle

\begin{abstract}
Long-horizon surgical robot manipulation is challenging because task
rewards are sparse, while meaningful interaction changes occur at
irregular intervals. Existing world-model agents typically imagine at
primitive-step resolution, leaving variable-duration task progress
implicit. Manually specified stages can provide intermediate structure,
but their task-specific boundaries are difficult to align with
state-dependent interaction transitions.

We propose S$^2$-HWM, a Sparse Event-Structured Hierarchical World Model
that learns sparse event evidence from primitive latent trajectories to
coordinate an event-level manager and a primitive-step worker. The event
evidence schedules manager goal updates, and each selected latent goal
conditions the worker's primitive actions until the next update. The learned
event evidence also
forms variable-duration segments for an Event Transition Model (ETM),
which predicts the next-boundary stochastic state, segment duration, and
accumulated segment reward. Chaining these event-level predictions
provides a variable-duration continuation beyond the primitive
imagination horizon for manager learning, while the worker retains
primitive-step actor--critic learning.

On a SurRoL-based PegTransfer task, S$^2$-HWM achieves a success rate of
$98.7\pm2.3\%$, outperforming the flat GASDreamerV3 baseline by 22.7
percentage points.
\end{abstract}

\section{Introduction}

Learning-based autonomous surgical manipulation seeks to improve the precision and efficiency of robotic task execution. 
Surgical manipulation often involves long-horizon tasks that require a sequence of coordinated actions and interactions over an extended period.
Even a basic transfer procedure may require the robot to approach an object, establish a stable grasp, transport it under geometric constraints, and release it at a target.
The timing of task transitions depends on the state of the evolving interaction.
However, intermediate progress boundaries are difficult to specify. 
Without clear boundaries, rewards that reflect partial task progress are difficult to assign, so learning signals are often available only at task completion. 
This sparse and delayed feedback makes reliable policy learning a central challenge in surgical robot learning.

Existing systems often alleviate this credit-assignment problem by introducing explicit intermediate task structure. 
This structure is often encoded through stage-dependent rewards, 
and provides useful intermediate signals for learning and decision making~\cite{riedmiller2018learning,huang2023viskill}. 
However, their temporal organization is fixed before execution and tailored to particular tasks. 
Coarse stages and unclear stage boundaries can easily lead to wrong interaction state assignments to the same phase.
Meanwhile, finer contact-dependent boundaries vary with execution and are difficult to define reliably in advance.

These boundaries should instead be inferred from the interaction dynamics experienced by the agent. 
World models provide a natural basis for this purpose. 
They encode interaction histories in latent states and predict how those states evolve under actions. 
Existing world-model agents use these learned dynamics to train policies through imagination \cite{hafner2019planet,hafner2020dreamer,hafner2021dreamerv2,hafner2023dreamerv3}. 
However, their transitions and imagined rollouts remain organized at primitive-step resolution. 
At this resolution, a variable-duration interaction is represented only as a sequence of fixed-step transitions, leaving its temporal boundaries implicit.
Consequently, they capture local evolution but do not explicitly represent when a meaningful event ends, how long it lasts, or how reward accumulates within it. 
As a result, the learned dynamics cannot directly organize high-level decisions or long-horizon credit assignment around variable-duration events.

This paper proposes S$^2$-HWM, a Sparse Event-Structured Hierarchical
World Model that learns event evidence from primitive latent
trajectories to organize prediction and control at two temporal scales.
The learned evidence supports event-aligned goal updates: the manager
selects a latent goal at each accepted update point, and the worker
executes primitive actions conditioned on this goal until the next
update. The event evidence is also used to construct variable-duration
training segments for an Event Transition Model (ETM), which predicts
the next-boundary stochastic state, segment duration, and accumulated
segment reward. Chaining these event-level predictions extends the
manager's value bootstrap beyond the primitive imagination horizon,
while the worker retains primitive-step actor--critic learning.

In summary, our main contributions are:
\begin{itemize}
    \item \textbf{Sparse event evidence for hierarchical control.}
    S$^2$-HWM learns sparse event evidence from primitive latent
    trajectories without semantic stage supervision. This evidence
    supports event-aligned manager goal updates and the
    construction of variable-duration event segments.

    \item \textbf{Event-level prediction for manager learning.}
    We introduce an ETM that predicts the next-boundary stochastic
    state, segment duration, and accumulated segment reward directly
    across variable-duration events. Chaining these predictions extends
    the manager's value bootstrap beyond the primitive imagination
    horizon without recursively predicting every intermediate primitive
    state.

    \item \textbf{Experimental evaluation.}
    We evaluate S$^2$-HWM on the SurRoL-based PegTransfer task against
    model-free and world-model baselines, and validate its key design
    choices through component ablations and event-level analyses.
\end{itemize}

\section{Related Work}

\subsection{Surgical Robot Learning}

Surgical robot learning requires autonomous agents to perform long-horizon manipulation under contact-rich dynamics, visual ambiguity, and staged task progression.
SurRoL offers a standardized dVRK-compatible simulation platform for surgical robot learning \cite{xu2021surrol}, and subsequent work has expanded the embodied simulation and task-autonomy ecosystem for laparoscopic robot-assisted surgery \cite{long2023human,yang2024softbody,long2025surgical}.
Within this domain, existing policy-learning methods expose or model temporal structure in different ways.
DEX uses demonstration-guided exploration for surgical robot task automation \cite{huang2023dex}.
ViSkill uses explicit subtask decomposition and scripted subtask demonstrations to learn value-informed skill chaining for long-horizon surgical tasks \cite{huang2023viskill}.
Goal-conditioned decision-transformer methods model surgical task execution as sequence prediction conditioned on goals and temporal indicators \cite{fu2024gcdt}.
Grasp Anything for Surgery (GAS) is the most closely related work to our visual-control setting, as it introduces a task-oriented visual abstraction and employs Dreamer-style world models to learn surgical grasping policies \cite{lin2024gas}.
Its evaluation focuses on GraspAny, where control is organized around a
comparatively short grasping sequence. S$^2$-HWM extends the GAS-style visual
abstraction and discrete action interface to PegTransfer, where grasping must
be followed by transport, alignment, and stable release, creating longer
temporal dependencies for policy learning.
Despite providing strong baselines and effective task abstractions, these approaches either rely on externally defined task structures, primarily focus on sequence modeling, or adopt flat world-model architectures for control, limiting their ability to capture hierarchical decision-making and long-horizon planning.
S$^2$-HWM learns sparse event evidence and event-level consequences to support
hierarchical credit assignment over these dependencies.

\subsection{World Models for Control}

World models learn predictive latent representations for planning, policy learning, or both.
PlaNet and Dreamer-style agents learn recurrent latent dynamics from observations and train policies through latent imagination \cite{hafner2019planet,hafner2020dreamer,hafner2021dreamerv2,hafner2023dreamerv3}.
TD-MPC2 represents a complementary control-centric direction, learning scalable implicit world models for continuous control and performing trajectory optimization in latent space \cite{hansen2024tdmpc2}.
These methods demonstrate the strength of learned world models for control, but their imagined trajectories are typically organized over fixed primitive steps or fixed planning horizons.
Several studies have incorporated hierarchical reasoning or temporal abstraction into world models.
Director learns latent high-level goals through imagined rollouts and optimizes a manager--worker policy based on world-model predictions \cite{hafner2022director}.
THICK introduces ContextRSSM, which uses sparse gated context updates to expose adaptive temporal boundaries, together with a high-level model that predicts event-end latent states, durations, and accumulated rewards across these boundaries \cite{thick2024}.
S$^2$-HWM couples event-structured prediction with hierarchical
manager--worker control by using the same event evidence to schedule
manager goal updates and construct ETM transitions. The ETM continuation
extends the manager's terminal bootstrap beyond primitive imagination,
while the worker continues to learn at primitive-step resolution.

\subsection{Hierarchical Reinforcement Learning}

Hierarchical reinforcement learning decomposes long-horizon control into high-level objectives and the primitive actions used to realize them.
This hierarchy requires specifying both what intermediate objective should guide the low-level policy and when that objective should be updated.
Existing methods represent high-level objectives as subgoals, intentions, or skills \cite{kulkarni2016hierarchical,vezhnevets2017feudal,riedmiller2018learning}.
Their temporal execution is commonly modeled using semi-Markov or option-based formulations, in which a selected objective remains active for a variable number of primitive steps \cite{sutton1999between}. 
The option-critic architecture further removes manually defined switching rules by learning both option policies and their termination conditions end-to-end \cite{bacon2017option}.
In robotic manipulation, however, high-level updates are still often governed by fixed intervals or manually specified subtask boundaries.
Such externally prescribed schedules may not track interaction changes that emerge during execution.
S$^2$-HWM instead learns sparse event evidence from latent trajectories and uses it to schedule event-aligned manager goal updates.

\begin{figure*}[t]
  \centering
  \includegraphics[width=\textwidth]{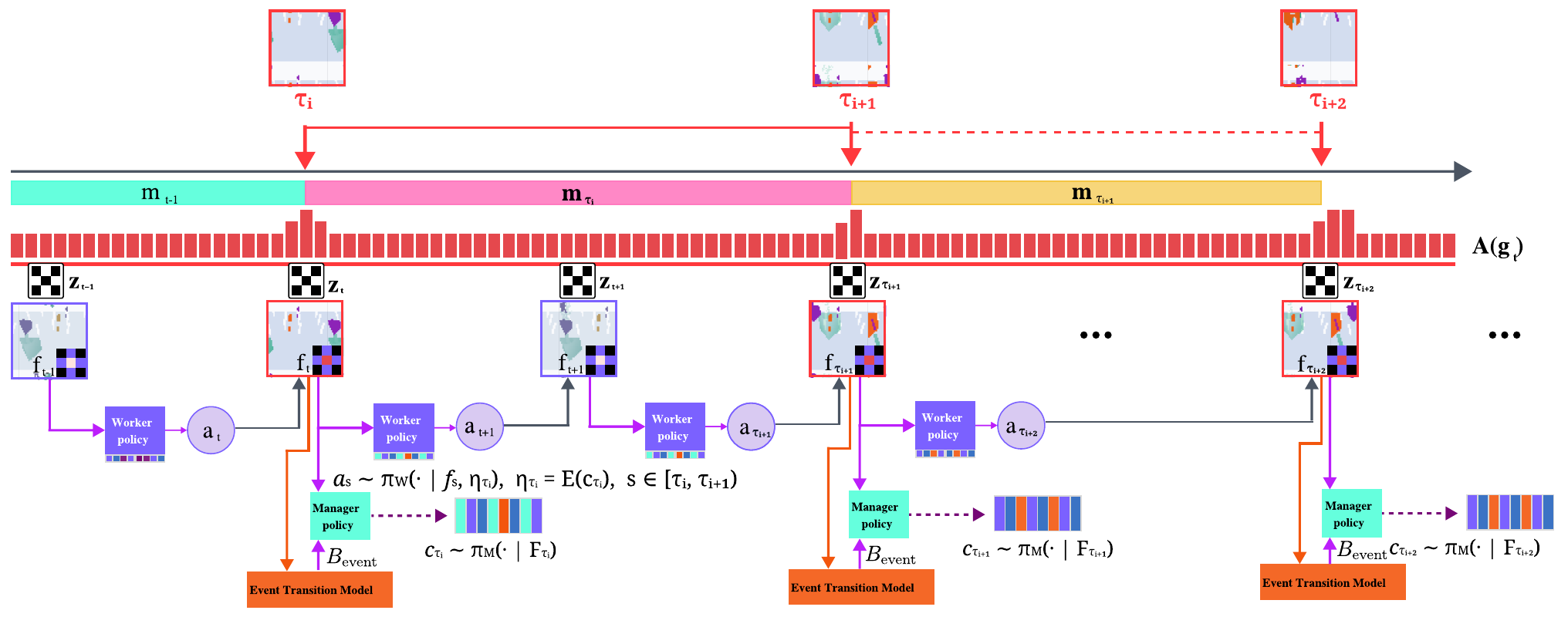}
  \caption{Event-aligned hierarchical execution in S$^2$-HWM. Primitive latent features $f_t$ support primitive-step worker actions, while $A(g_t)$ denotes the manager-side acceptance rule that maps sparse event evidence $g_t$ to accepted manager update points $\tau_i$. At each accepted manager update point, the manager samples a discrete code $c_i$ from $F_i=f_{\tau_i}$, and its embedding $\eta_i$ conditions the worker until the next manager update. Dashed ETM paths denote the event-level continuation used in the manager bootstrap; its construction is detailed in Figure~\ref{fig:two-timescale-imagination}.}
  \label{fig:s2hwm-overview}
\end{figure*}

\section{Method}
This paper proposes S$^2$-HWM, a Sparse Event-Structured Hierarchical
World Model that organizes primitive-step latent dynamics through event
evidence learned directly from latent trajectories.
The learned evidence identifies changes in interaction context and
supports event-aligned manager goal updates.
At each accepted manager update point, the manager selects a latent goal
from the current latent feature, and this goal conditions the worker's
primitive actions until the next update, forming the hierarchical
execution process illustrated in Figure~\ref{fig:s2hwm-overview}.
For event-level learning, a separate acceptance rule is applied to the
same event evidence to construct variable-duration training segments,
over which the Event Transition Model (ETM) predicts the next-boundary
stochastic state, segment duration, and accumulated segment reward.
The learned event evidence therefore provides a shared signal for
manager scheduling and ETM training, while chained ETM predictions
extend the manager's value bootstrap beyond the primitive imagination
horizon.

\subsection{Event-Aligned Hierarchical Policy Learning}

Figure~\ref{fig:s2hwm-overview} illustrates the hierarchical policy
operating at two temporal scales.
At every primitive step, the worker samples an action conditioned on the
current latent feature $f_t$ and the active manager goal $\eta_i$.
At an accepted manager update point $\tau_i$, the manager observes the boundary feature
$F_i=f_{\tau_i}$, samples a discrete code $c_i$, and maps it to a new
latent goal through the learned codebook $E$.
This goal remains active until the next update and conditions all
worker actions within the corresponding variable-duration segment:
\begin{equation}
\begin{aligned}
c_i&\sim\pi_M(\cdot\mid F_i),
\qquad \eta_i=E(c_i),\\
a_t&\sim\pi_W(\cdot\mid f_t,\eta_i),
\qquad t\in[\tau_i,\tau_{i+1}).
\end{aligned}
\label{eq:hierarchical-policy}
\end{equation}

The policy in Eq.~\eqref{eq:hierarchical-policy} requires a latent
representation that is updated at every primitive step, summarizes
interaction history under partial observability, and supports imagined
control.
We therefore instantiate the primitive world model as a DreamerV3
recurrent state-space model (RSSM) \cite{hafner2023dreamerv3}.
Given the observation--action history, the RSSM updates the deterministic
state $h_t$ as recurrent memory and infers the stochastic state $z_t$ to
represent the current latent interaction state.
Together with the persistent interaction context $m_t$, these states
form
\begin{equation}
\bar f_t=[z_t,h_t],
\qquad
f_t=[\bar f_t,m_t]=[z_t,h_t,m_t].
\label{eq:context-augmented-feature}
\end{equation}
The evolving feature $f_t$ provides the history-dependent state required
for primitive worker control, while its boundary snapshot
$F_i=f_{\tau_i}$ provides the manager with the representation used for
latent-goal selection.
The selected action $a_t$ is supplied to the next RSSM transition,
closing the primitive-step control loop.

The event-aligned update points required by the manager are derived from
the sparse context dynamics described next.

\subsection{Sparse Context Dynamics and Event Boundaries}

The manager update schedule must reflect interaction changes that depend
on the preceding observation--action history rather than a single
observation. S$^2$-HWM formulates surgical manipulation as a partially
observable Markov decision process (POMDP), for which the RSSM infers a
history-dependent latent state. Although this state supports
primitive worker control, it evolves at every step, so its local
variations do not reliably indicate when the manager should refresh its
goal. We therefore augment it with a persistent interaction context
$m_t$, whose sparse revisions provide evidence of event-level changes.

Given the previous stochastic state, action, and context, the context
model predicts a candidate update $\tilde m_t$ and a soft gate $\rho_t$:
\begin{equation}
\begin{aligned}
(\tilde m_t,\rho_t)
&=\Phi_{\mathrm{ctx}}(z_{t-1},a_{t-1},m_{t-1}),
\qquad \rho_t\in[0,1]^d,\\
m_t
&=\rho_t\odot\tilde m_t
 +(1-\rho_t)\odot m_{t-1},\\
g_t&=\operatorname{ST}(\rho_t).
\end{aligned}
\label{eq:sparse-context-update}
\end{equation}

The candidate \(\tilde m_t\) proposes new context information, while
\(\rho_t\) controls how much of it is incorporated into \(m_t\).
Straight-through binarization produces the sparse dimension-wise
activation \(g_t\), while regularization limits excessive updates. The
resulting \(g_t\) serves as event evidence rather than a temporal boundary
indicator; accepted boundaries are obtained by aggregating these
activations and applying the corresponding acceptance rule.

For notational simplicity, we use \(\{\tau_i\}\) to denote the accepted
boundary sequence in each context.
Manager scheduling and ETM segment construction apply separate
acceptance rules to the same event evidence \(g_t\); therefore, their
accepted boundaries need not coincide exactly. In the ETM context,
adjacent accepted boundaries define
\begin{equation}
\begin{aligned}
e_i &= [\tau_i,\tau_{i+1}), &
\Delta_i &= \tau_{i+1}-\tau_i, &
x_i^e &= [z_{\tau_i},m_{\tau_i}].
\end{aligned}
\label{eq:event-segments}
\end{equation}
The resulting variable-duration segments adapt to the interaction
trajectory without semantic stage annotations.

\subsection{Event Transition Model}
To extend manager learning beyond the primitive imagination horizon, the
ETM models transitions between accepted boundaries at event resolution.
Each transition spans a variable-duration segment and predicts the
next-boundary stochastic state, segment duration, and accumulated segment
reward without recursively rolling out the intermediate primitive states.
This reduces the number of recursive model transitions required to cover
the same amount of physical time and limits the opportunities for
prediction errors to compound.

\begin{figure}[!t]
  \centering
  \includegraphics[width=\columnwidth]{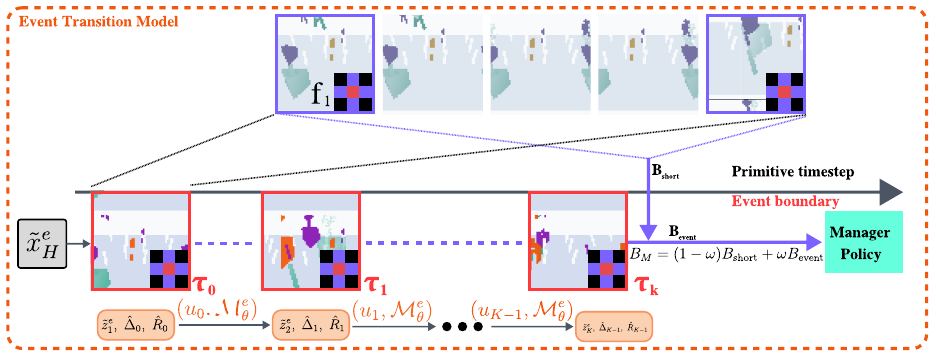}
  \caption{Two-timescale imagination and event-extended manager bootstrap.
  Primitive-step imagination produces the short-horizon terminal estimate
  $B_{\mathrm{short}}$. Starting from the same endpoint, $K$ ETM transitions
  predict next-boundary stochastic states, segment durations, and accumulated
  segment rewards to form the event-level continuation $B_{\mathrm{event}}$.
  Their weighted combination $B_M$ extends the manager value target while
  worker learning remains at primitive-step resolution. Boundary observations are
  shown only for visualization; the ETM predicts latent states rather than
  pixels.}
  \label{fig:two-timescale-imagination}
\end{figure}

\paragraph{Event Transition Learning.}
We train the ETM on latent trajectory segments between consecutive
accepted ETM boundaries. For \(e_i=[\tau_i,\tau_{i+1})\), the ETM takes
the event state \(x_i^e\) at the starting boundary as input and predicts
the stochastic state \(z_{i+1}^e=z_{\tau_{i+1}}\) at the next boundary,
the segment duration
\(\Delta_i=\tau_{i+1}-\tau_i\), and the accumulated segment reward
\(R_i=\sum_{k=0}^{\Delta_i-1}r_{\tau_i+k}\).

Because the same starting event state may be followed by different
next-boundary outcomes, we introduce an internal categorical transition
mode \(u_i\) to represent the variability among observed transitions.
During training, the posterior
\(q_\theta(u_i\mid x_i^e,z_{i+1}^e)\) infers the transition mode using
the observed next-boundary stochastic state, whereas the prior
\(p_\theta(u_i\mid x_i^e)\) predicts it without future information.
The transition mode \(u_i\) is internal to the ETM and is distinct from
the manager code \(c_i\). Conditioned on the event state and transition
mode, the ETM predicts
\begin{equation}
(\hat z_{i+1}^{e},\hat\Delta_i,\hat R_i)
=
\mathcal M_{\theta}^{e}(x_i^{e},u_i).
\label{eq:etm-transition}
\end{equation}
A KL regularizer aligns the posterior with the prior. During imagination,
the most probable mode under the prior conditions each ETM transition.

\paragraph{Event-Extended Manager Bootstrap.}
The ETM predictions contribute to hierarchical policy learning by
extending the terminal bootstrap used for the manager target. As
illustrated in Figure~\ref{fig:two-timescale-imagination}, this target
combines a short-horizon estimate from primitive-step imagination with
a continuation over future event transitions.

In the upper branch, the RSSM and hierarchical policy generate a
primitive-step imagined trajectory \(\tilde f_{0:H}\). The manager value
at its terminal state defines the short-horizon bootstrap
\(B_{\mathrm{short}}=V_M(\tilde f_H)\). This rollout preserves the local
dynamics required for worker learning, but its fixed horizon covers only
a limited portion of future task progress.

The lower branch continues from \(\tilde f_H\) using \(K\) recursive ETM
transitions. At each transition, the ETM predicts the next-boundary
stochastic state, segment duration, and accumulated segment reward. The
predicted stochastic state, together with the retained context,
initializes the next event state. The terminal RSSM recurrent state is
also retained when constructing the corresponding feature for manager
value evaluation. These predicted event states provide future latent
features for evaluating the manager value function and need not coincide
with the manager's accepted goal-update points. Accumulating the predicted
segment rewards with duration-dependent discounting
\(\gamma^{\hat\Delta_j}\) and bootstrapping from the final predicted
feature produces the event-level continuation \(B_{\mathrm{event}}\).

The final manager bootstrap combines the primitive estimate with
the event-level continuation:
\begin{equation}
B_M=(1-\omega)B_{\mathrm{short}}+\omega B_{\mathrm{event}},
\label{eq:manager-bootstrap}
\end{equation}
where \(\omega\) controls the event-level contribution. The resulting
\(B_M\) extends the manager target beyond the primitive imagination
horizon, while the worker retains its primitive-step actor--critic
target.

\begin{figure*}[!t]
\centering
\includegraphics[width=\textwidth]{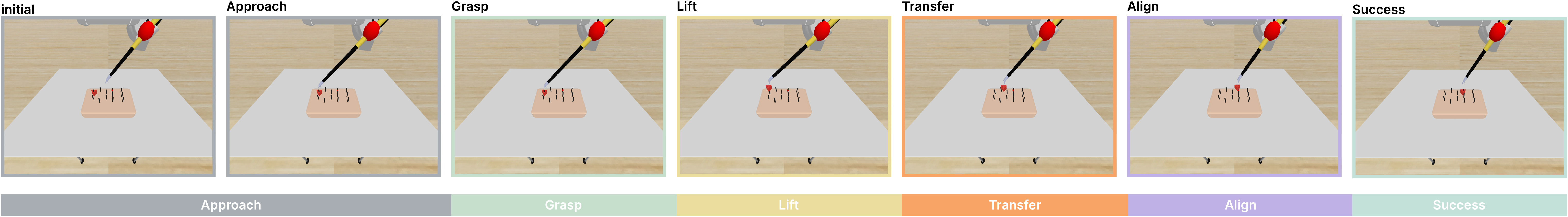}
\caption{Representative PegTransfer progress in SurRoL. Post-hoc semantic phase labels are used only for evaluation and visualization.}
\label{fig:pegtransfer_rollout}
\end{figure*}

\begin{figure*}[!t]
\centering
\includegraphics[width=\textwidth]{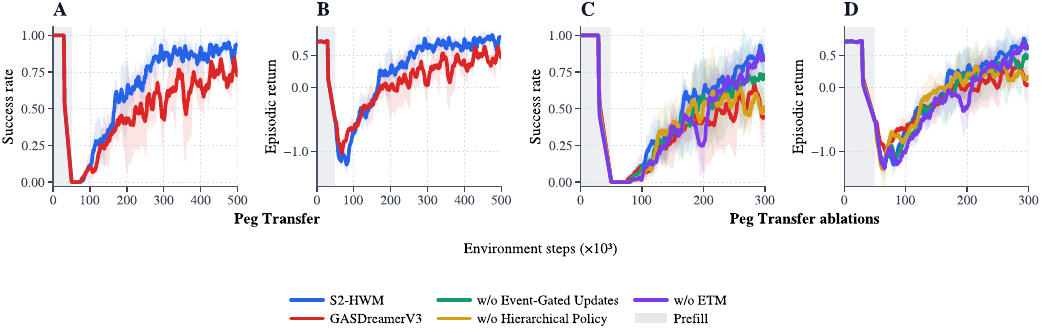}
\caption{PegTransfer training performance and ablations. Panels A and B compare
S$^2$-HWM with GASDreamerV3 through 500k environment steps. Panels C and D
compare S$^2$-HWM with \emph{w/o Event-Gated Updates} ($K=15$),
\emph{w/o ETM}, and \emph{w/o Hierarchical Policy} over a common 300k-step
window. Lines
show means, and shaded bands show sample standard deviations across three
training seeds. Gray denotes prefill.}
\label{fig:task_performance}
\end{figure*}

\section{Experiments}

\subsection{Experimental Setup}

To evaluate long-horizon sparse-reward surgical manipulation, all
compared methods are trained and evaluated on a randomized SurRoL-based
PegTransfer task \cite{xu2021surrol}. Successful execution requires the
robot to approach and grasp a block, transport it under geometric
constraints, align it with the target peg, and release it in a stable
configuration. An episode is counted as successful when the released
block is stably placed on the target peg.
Figure~\ref{fig:pegtransfer_rollout} illustrates this progression using
post-hoc semantic labels that are not provided to the agent.

The comparison includes PPO \cite{schulman2017proximal}, DreamerV2
\cite{hafner2021dreamerv2}, and GASDreamerV3 \cite{lin2024gas}.
For the primary controlled comparison, GASDreamerV3 is retrained from its
original GraspAny setting on PegTransfer.
Following GAS \cite{lin2024gas}, S$^2$-HWM and GASDreamerV3 use the same
DSA4-depth observation frontend, task randomization, action interface, and
evaluation protocol, forming the primary controlled comparison. PPO and
DreamerV2 do not use DSA and serve
as additional model-free and earlier world-model references. All methods
share the SurRoL task interface, initial-state randomization, and nine-way
discrete action space; environment-step budgets are matched within each
comparison. For S$^2$-HWM and GASDreamerV3, the DSA4-depth frontend produces
four \(64\times64\) task-oriented channels. The action space contains six
Cartesian translations, two yaw rotations, and one gripper-toggle action.

All methods use three training seeds. Deterministic checkpoint
evaluation uses 50 episodes per seed, with success rates reported as
the mean and sample standard deviation across seeds.

The same checkpoints are further evaluated in two extended-horizon stress
tests. \emph{Repeated transfer} requires two consecutive transfers, while
\emph{repeated transfer with perturbation} introduces an external drop during
the second transfer. Single-transfer episodes allow 300 primitive steps; both
stress tests allow up to 800 steps and require both transfers to be completed.

Three matched-budget ablations examine the proposed components.
\emph{w/o Event-Gated Updates} retains the sparse context dynamics, ETM, and
manager--worker hierarchy but updates the manager every $K=15$ primitive
steps. \emph{w/o ETM} retains event-evidence-based manager updates while
removing ETM prediction and $B_{\mathrm{event}}$. \emph{w/o Hierarchical
Policy} retains the sparse context dynamics but replaces the
manager--worker policy with a flat actor.

\subsection{Task Performance and Ablations}

\begin{table*}[!t]
\centering
{\small
\begin{tabular*}{\textwidth}{@{\extracolsep{\fill}}lccc}
\toprule
Method
& Single transfer
& Repeated transfer
& Repeated transfer with perturbation \\
\midrule
PPO & $0.0 \pm 0.0$ & $0.0 \pm 0.0$ & $0.0 \pm 0.0$ \\
DreamerV2 & $0.0 \pm 0.0$ & $0.0 \pm 0.0$ & $0.0 \pm 0.0$ \\
GASDreamerV3 & $76.0 \pm 12.0$ & $77.3 \pm 18.6$ & $65.3 \pm 14.0$ \\
S$^2$-HWM
& $\mathbf{98.7 \pm 2.3}$
& $\mathbf{91.3 \pm 6.1}$
& $\mathbf{88.0 \pm 5.3}$ \\
\midrule
w/o Event-Gated Updates
& $80.7 \pm 28.3$
& $64.7 \pm 56.1$
& $63.3 \pm 55.1$ \\
w/o ETM
& $91.3 \pm 5.0$
& $74.7 \pm 38.7$
& $67.3 \pm 30.0$ \\
w/o Hierarchical Policy
& $68.0 \pm 20.3$
& $52.0 \pm 33.4$
& $46.0 \pm 30.0$ \\
\bottomrule
\end{tabular*}
}
\caption{Deterministic PegTransfer performance across nominal and
extended-horizon evaluation settings. All settings use matched checkpoints
at approximately 300k environment steps. Repeated transfer requires two
sequential transfers within one episode, while repeated transfer with
perturbation externally interrupts the second transfer with an induced drop.
All entries report SR (\%); values are mean $\pm$ sample standard deviation
across three training seeds, each evaluated over 50 episodes. PPO and DreamerV2
obtain zero success in all three settings.}
\label{tab:task_success}
\end{table*}

Figure~\ref{fig:task_performance} and Table~\ref{tab:task_success}
evaluate S$^2$-HWM through its training dynamics, matched-checkpoint
performance, and component ablations.

Following the common prefill period, both S$^2$-HWM and GASDreamerV3
begin to acquire successful behavior, after which their learning curves
gradually separate. S$^2$-HWM improves more rapidly and maintains higher
mean success and episodic return throughout the later training stage.
At the matched checkpoint, S$^2$-HWM achieves $98.7\pm2.3\%$ SR,
compared with $76.0\pm12.0\%$ for GASDreamerV3, a difference of 22.7
percentage points. PPO and DreamerV2 do not solve the task within the
same training budget. Because S$^2$-HWM and GASDreamerV3 share the DSA
frontend, action interface, and a DreamerV3-based primitive world-model
backbone, their comparison evaluates the overall benefit of introducing
event organization and hierarchical policy learning over a flat
world-model agent.

The ablations first show that learning a sparse event representation
alone is insufficient for effective control. Replacing the
manager--worker policy with a flat actor while retaining the sparse
context dynamics reduces single-transfer SR to $68.0\pm20.3\%$, the
largest degradation among the three variants. The learned event evidence
must therefore be converted by the manager into persistent latent goals
and realized through the worker's primitive actions. The hierarchy
provides this connection between event-level decisions and
primitive-step control.

Manager update timing also contributes beyond the hierarchical
architecture itself. Retaining the manager--worker policy, sparse
context dynamics, and ETM but replacing event-aligned updates with a
fixed $K=15$ schedule reduces SR to $80.7\pm28.3\%$ and substantially
increases variation across seeds. Because this variant changes only the
manager update rule, the result suggests that event evidence is useful
not only as a latent representation but also for deciding when the
high-level goal should be refreshed. A single fixed period cannot adapt
to the different durations of approach, grasp, transport, and placement
interactions.

The ETM provides a smaller but measurable additional benefit on the
nominal single-transfer task. Removing ETM continuation while retaining
event-aligned manager updates and the hierarchical policy yields
$91.3\pm5.0\%$ SR, 7.4 percentage points below the complete model.
Thus, event-aligned hierarchical control already provides substantial
task structure, while the ETM further improves manager learning by
extending its bootstrap beyond the primitive imagination horizon.

Without additional training, the same checkpoints are then evaluated
on both stress tests. On repeated transfer, S$^2$-HWM achieves
$91.3\pm6.1\%$ SR versus $77.3\pm18.6\%$ for GASDreamerV3.
When a drop interrupts the second transfer, S$^2$-HWM retains
$88.0\pm5.3\%$ SR versus $65.3\pm14.0\%$ for GASDreamerV3 and
outperforms every ablation in both settings.

The extended settings make the ETM contribution more pronounced.
Removing the ETM causes a 7.4-point reduction on single transfer, but
the gap increases to 16.6 points on repeated transfer and 20.7 points
under perturbation. This pattern is consistent with the ETM design:
when execution spans additional events or must continue after an
interruption, primitive-step imagination covers a smaller fraction of
the remaining task, and event-level continuation provides the manager
with a more distant value estimate. Overall, the hierarchy converts
event information into control, event-aligned updates determine when
the high-level goal changes, and the ETM extends the future task
progress evaluated by the manager. The following subsection examines
the corresponding predictive and behavioral structure.

\begin{figure}[!t]
\centering
\includegraphics[width=\columnwidth]{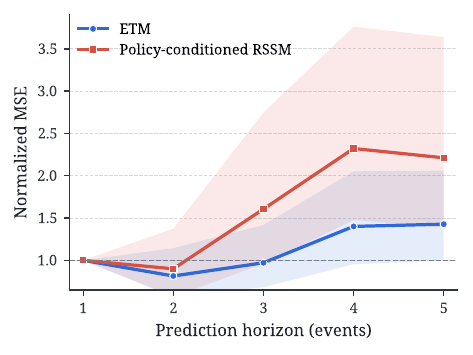}
\caption{Normalized multi-event stochastic-state prediction error on
PegTransfer. ETM and
policy-conditioned RSSM rollouts start from the same posterior stochastic states at event boundaries
and are evaluated at matching future boundaries. Each curve is normalized by its
own one-event MSE; lines show means and shaded regions show trajectory-level
dispersion. Lower is better.}
\label{fig:multievent_prediction}
\end{figure}

\begin{table*}[!t]
\centering
{\small
\begin{tabular*}{\textwidth}{@{\extracolsep{\fill}}lccccccccc}
\toprule
Post-hoc semantic phase & $x$- & $x$+ & $y$- & $y$+ & $z$- & $z$+ & yaw- & yaw+ & grip \\
\midrule
Approach-source & 13\% & \actsecondary{18\%} & 15\% & 11\% & \actprimary{25\%} & 8\% & 3\% & 2\% & 5\% \\
Approach-grasp & 7\% & 0\% & 1\% & 6\% & \actprimary{30\%} & \actsecondary{27\%} & 0\% & 0\% & \actsecondary{29\%} \\
Grasp-lift & \actsecondary{16\%} & 10\% & 1\% & 13\% & 13\% & \actprimary{41\%} & 0\% & 0\% & 6\% \\
Place-align & \actprimary{41\%} & 2\% & 4\% & \actsecondary{38\%} & 8\% & 3\% & 1\% & 0\% & 3\% \\
Release-stable & \actprimary{27\%} & 8\% & 10\% & 13\% & \actsecondary{15\%} & 10\% & 3\% & 2\% & 12\% \\
\bottomrule
\end{tabular*}
}
\caption{Worker action occurrence ratios on deterministic PegTransfer
evaluation rollouts from a representative S$^2$-HWM seed-0 policy. Each manager
code is assigned post hoc to its dominant semantic phase: Approach-source
$\{07,10,18,23,28,56,58\}$; Approach-grasp $\{00,31\}$; Grasp-lift
$\{20,35,36\}$; Place-align $\{49,60\}$; and Release-stable
$\{03,04,22,33,38,43,45,47,54\}$. Rows aggregate primitive steps from these code
sets, and entries are percentages over the nine primitive actions. Bold entries
mark the dominant preferences in each row.}
\label{tab:code_action_stage}
\medskip

\centering
\includegraphics[width=\textwidth]{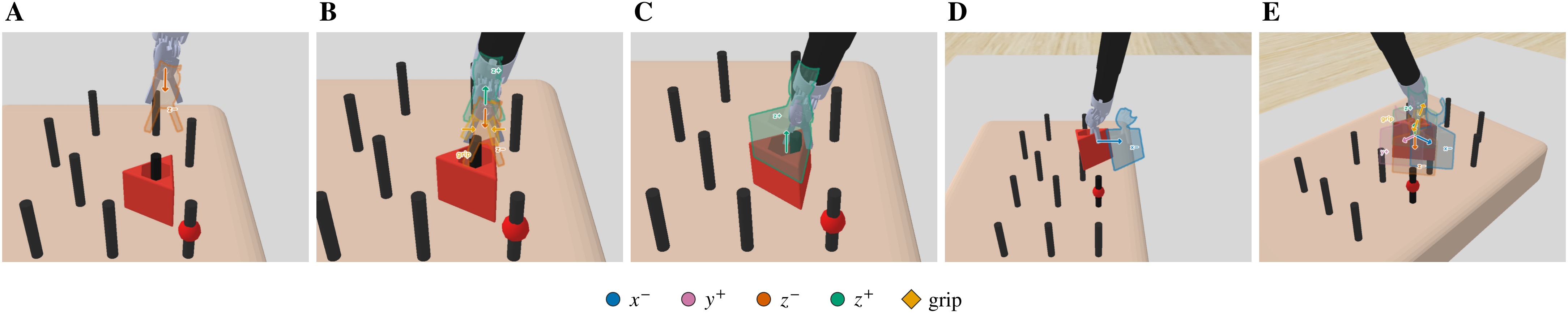}
\captionof{figure}{Worker action tendencies for the post-hoc semantic phases in
Table~\ref{tab:code_action_stage}: (A) Approach-source, (B) Approach-grasp,
(C) Grasp-lift, (D) Place-align, and (E) Release-stable. Colored afterimages and
arrows denote empirical world-frame actions, with opacity reflecting their
occurrence ratios. They visualize executed actions rather than predicted
trajectories.}
\label{fig:stage_action_tendencies}
\end{table*}

\subsection{Event Prediction and Policy Analysis}

Task success does not directly reveal whether the event-level mechanisms
behave as intended. We therefore analyze ETM prediction error across multiple
future events and the relationship between manager codes and worker behavior.

For multi-event prediction, ETM and policy-conditioned RSSM rollouts start
from the same posterior stochastic states at event boundaries and are evaluated against posterior
stochastic states at matching future boundaries. Each model's error is
normalized by its own one-event MSE, so the metric measures relative error
growth rather than absolute one-step accuracy. Figure~\ref{fig:multievent_prediction}
shows that ETM error remains close to the one-event reference through three
event transitions and subsequently grows more slowly than RSSM error. Although
the uncertainty bands overlap at longer horizons, the mean trend supports
using event-level continuation to extend the manager bootstrap beyond primitive
imagination.

For policy analysis, manager codes from deterministic rollouts of the
representative seed-0 policy are grouped post hoc according to the semantic phase
in which they are most frequently active. Table~\ref{tab:code_action_stage}
and Figure~\ref{fig:stage_action_tendencies} show that the resulting groups
exhibit different primitive-action tendencies. Vertical adjustment and the
gripper action are prominent near grasping, upward motion dominates
grasp--lift, and lateral corrections characterize placement. The labels and
group assignments are not used during training; these results therefore do
not imply that manager codes reproduce semantic phases, but show that
manager-conditioned intervals are associated with structured worker behavior.

\FloatBarrier
\section{Conclusion}

We presented S$^2$-HWM, a Sparse Event-Structured Hierarchical World
Model for long-horizon sparse-reward surgical manipulation. Its central
idea is to use learned event evidence as a shared interface between
primitive-step control and event-level decision making. The event
evidence schedules manager goal updates and constructs variable-duration
segments for the ETM, which predicts boundary-to-boundary outcomes to
extend the manager bootstrap beyond the primitive imagination horizon.
On the SurRoL-based PegTransfer task, S$^2$-HWM achieves
$98.7\pm2.3\%$ success, outperforming GASDreamerV3 by 22.7 percentage
points, and retains its advantage in repeated-transfer and perturbation
stress tests. Matched-budget ablations support the contributions of the
hierarchy, event-aligned updates, and ETM continuation, while diagnostic
analyses associate the learned organization with interaction-dependent
transitions and structured worker behavior without semantic stage
supervision. These findings support learned event evidence as an
effective connection between world modeling and hierarchical control.
The current evaluation is limited to a simulated PegTransfer task
family; future work will extend the framework to broader surgical tasks
and physical robotic systems.

\bibliography{references}

\end{document}

%% file: author_info.tex
\newcommand{\ArxivPDFAuthors}{Shuzhe Zhang; Xin Zhu; Yinling Qian; Qiong Wang}

\author{
Shuzhe Zhang\textsuperscript{\rm 1},
Xin Zhu\textsuperscript{\rm 1},
Yinling Qian\textsuperscript{\rm 1,2}\corresponding,
Qiong Wang\textsuperscript{\rm 1}
}
\affiliations{
\textsuperscript{\rm 1}Shenzhen Institutes of Advanced Technology, Chinese Academy of Sciences\\
\textsuperscript{\rm 2}University of Chinese Academy of Sciences
}